\documentclass[10pt,twocolumn,letterpaper]{article}
\pdfoutput=1
\usepackage{cvpr}
\usepackage{times}
\usepackage{graphicx}
\usepackage{amsmath}
\usepackage{pifont}
\usepackage{amssymb}
\usepackage{caption}
\usepackage{color}
\usepackage{algpseudocode}
\usepackage{multirow}
\usepackage{wrapfig}
\usepackage{lipsum}
\usepackage{enumitem}
\usepackage{rotating}
\usepackage[export]{adjustbox}
\usepackage[ruled]{algorithm2e}
\usepackage{pythontex}

\newcommand{\model}{TTNet\xspace}

\cvprfinalcopy 

\def\cvprPaperID{5230} 

\ifcvprfinal\fi
\begin{document}

\title{Zero-Shot Task Transfer}

\author{Arghya Pal, Vineeth N Balasubramanian\\
Department of Computer Science and Engineering\\
Indian Institute of Technology, Hyderabad, INDIA\\
{\tt\small \{cs15resch11001, vineethnb\}@iith.ac.in}
}

\maketitle
\begin{abstract}
In this work, we present a novel meta-learning algorithm \model \footnote{This work is accepted as an Oral presentation in Computer Vision and Pattern Recognition CVPR 2019} that regresses model parameters for novel tasks for which no ground truth is available (zero-shot tasks). In order to adapt to novel zero-shot tasks, our meta-learner learns from the model parameters of known tasks (with ground truth) and the correlation of known tasks to zero-shot tasks. Such intuition finds its foothold in cognitive science, where a subject (human baby) can adapt to a novel concept (depth understanding) by correlating it with old concepts (hand movement or self-motion), without receiving an explicit supervision. We evaluated our model on the Taskonomy dataset, with four tasks as zero-shot: surface normal, room layout, depth and camera pose estimation. These tasks were chosen based on the data acquisition complexity and the complexity associated with the learning process using a deep network. Our proposed methodology outperforms state-of-the-art models (which use ground truth) on each of our zero-shot tasks, showing promise on zero-shot task transfer. We also conducted extensive experiments to study the various choices of our methodology, as well as showed how the proposed method can also be used in transfer learning. To the best of our knowledge, this is the first such effort on zero-shot learning in the task space.
\end{abstract}
\section{Introduction}
The major driving force behind modern computer vision, machine learning, and deep neural network models is the availability of large amounts of curated labeled data. Deep models have shown state-of-the-art performances on different vision tasks. Effective models that work in practice entail a requirement of very large labeled data due to their large parameter spaces. Expecting availability of large-scale hand-annotated datasets for every vision task is not practical. Some tasks require extensive domain expertise, long hours of human labor, expensive data collection sensors - which collectively make the overall process very expensive. Even when data annotation is carried out using crowdsourcing (e.g. Amazon Mechanical Turk), additional effort is required to measure the correctness (or goodness) of the obtained labels. Due to this, many vision tasks are considered expensive \cite{zamir2016generic}, and practitioners either avoid such tasks or continue with lesser amounts of data that can lead to poorly performing models. We seek to address this problem in this work, viz., to build an alternative approach that can obtain model parameters for tasks without any labeled data. Extending the definition of zero-shot learning from basic recognition settings, we call our work \textit{Zero-Shot Task Transfer}.
\begin{figure}
 \centering 
 \includegraphics[width=7.8cm, height=4.2cm]{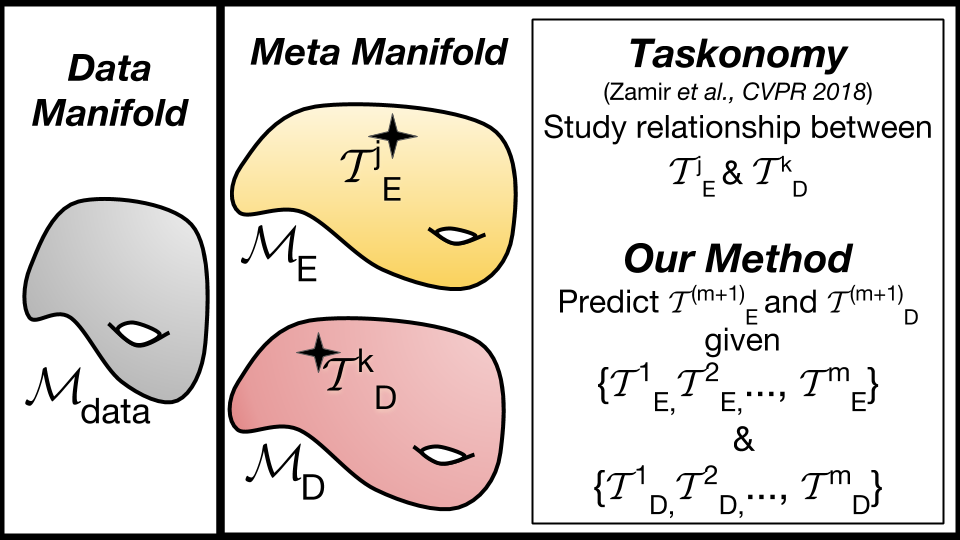}
 \caption{Our \textbf{Zero-Shot Task Transfer} framework explores meta-manifold of model parameters to regress model parameters of zero-shot tasks for which no ground truth is available. We compare our objective with that of Taskonomy \cite{zamir2018taskonomy} to delineate the difference. Our algorithm \model (described in section \ref{sec_methodology} assume data manifold $\mathcal{M}_{data}$ and meta-manifold that is furthur divided into meta-encoder manifold $\mathcal{M}_E$, and meta-decoder manifold $\mathcal{M}_D$.}
 \label{fig_1_figure_overview}
\end{figure}

Cognitive studies show results where a subject (human baby) can adapt to a novel concept (e.g. depth understanding) by correlating it with known concepts (hand movement or self-motion), without receiving an explicit supervision. In similar spirit, we present our meta-learning algorithm that computes model parameters for novel tasks for which no ground truth is available (called \textit{zero-shot tasks}). In order to adapt to a zero-shot task, our meta-learner learns from the model parameters of known tasks (with ground truth) and their task correlation to the novel task. Formally, given the knowledge of $m$ known tasks \{$\tau_1, \cdots, \tau_m$\}, a meta-learner $\mathcal{F(.)}$ can be used to extrapolate parameters for $\tau_{(m+1)}$, a novel task. 
 
However, with no knowledge of relationships between the tasks, it may not be plausible to learn a meta-learner, as its output could map to any point on the meta-manifold (see Figure \ref{fig_1_figure_overview}). We hence consider the task correlation between known tasks and a novel task as an additional input to our framework. There could be different notions on how task correlation is obtained. In this work, we use the approach of wisdom-of-crowd for this purpose. Many vision \cite{pal2018adversarial} and non-vision machine learning applications \cite{ratner2016data}, \cite{varma2016socratic} encode such crowd wisdom in their learning methods. Harvesting task correlation knowledge from the crowd is fast, cheap, and brings domain knowledge. High-fidelity aggregation of crowd votes is used to integrate the task correlation between known and zero-shot tasks in our model. We however note that our framework can admit any other source of task correlation beyond crowdsourcing. (We show our results with other sources in the supplementary section.) 

Our broad idea of leveraging task correlation can be found similar to the recently proposed idea of Taskonomy \cite{zamir2018taskonomy}, but our method and objectives are different in many ways (see Figure \ref{fig_1_figure_overview}): (i) Taskonomy studies task correlation to find a way to transfer one task model to another, while our method extrapolates to a zero-shot task, for which no labeled data is available; (ii) To adapt to a new task, Taskonomy requires a considerable amount of target labeled data, while our work does not require any target labeled data (which is, in fact, our objective); (iii) Taskonomy obtains a task transfer graph based on the representations learned by neural networks; while in this work, we leverage task correlation to learn new tasks; and (iv) Lastly, our method can be used to learn multiple novel tasks simultaneously. As stated earlier, though we use crowdsourced task correlation, any other compact notion of task correlation can easily be encoded in our methodology. More precisely, our proposal in this work is not to learn an optimal task relation, but to extrapolate to zero-shot tasks. 

Our contributions can be summarized as follows:
\begin{itemize}[noitemsep,topsep=0pt]
\item We propose a novel methodology to infer zero-shot task parameters that be used to solve vision tasks with no labeled data.
\item The methodology can scale to solving multiple zero-shot tasks simultaneously, as shown in our experiments. Our methodology provides near state-of-the-art results by considering a smaller set of known tasks, and outperforms state-of-the-art models (learned with ground truth) when using all the known tasks, although trained with no labeled data.
\item We also show how our method can be used in a transfer learning setting, as well as conduct various studies to study the effectiveness of the proposed method.
\end{itemize}
\begin{figure*}
 \centering 
 \includegraphics[width=16cm, height=4.7cm]{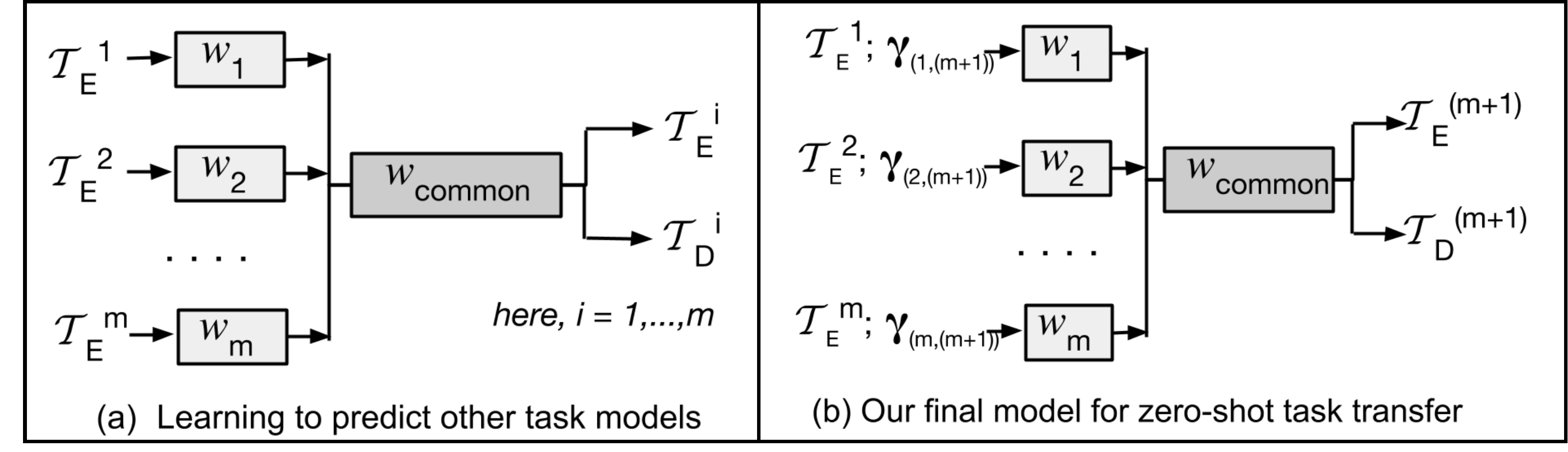}
 \caption{Overview of our work}
 \label{fig_2_final_figure_manifold}
\end{figure*}
\section{Related Work}
\label{sec_related_work}
We divide our discussion of related work into subsections that capture earlier efforts that are related to ours from different perspectives.
\paragraph{Transfer Learning:} Reusing supervision is the core component of transfer learning, where an already learned model of a task is finetuned to a target task. From the early experimentation on CNN features \cite{yosinski2014transferable}, it was clear that initial layers of deep networks learn similar kind of filters, can can hence be shared across tasks. Methods such as in \cite{aytar2011tabula}, \cite{lim2011transfer} augment generation of samples by transferring knowledge from one category to another. Recent efforts have shown the capability to transfer knowledge from model of one task to a completely new task \cite{ren2015faster}\cite{redmon2016you}. Zamir \textit{et al.} \cite{zamir2018taskonomy} extended this idea and built a task graph for 26 vision tasks to facilitate task transfer. However, unlike our work, \cite{zamir2018taskonomy} cannot be generalized to a novel task without accessing the ground truth.
\vspace{-0.5cm}
\paragraph{Multi-task Learning:} Multi-task learning learns multiple tasks simultaneously with a view of task generalization. Some methods in multi-task learning assume a prior and then iterate to learn a joint space of tasks \cite{cohen2014learning}\cite{kumar2012learning}, while other methods \cite{long2017learning}\cite{kumar2012learning} do not use a prior but learn a joint space of tasks during the process of learning. Distributed multi-task learning methods \cite{liu2017distributed} address the same objective when tasks are distributed across a network. However, unlike our method, a binding thread for all these methods is that there is an explicit need of having labeled data for all tasks in the setup. These methods can not solve a zero-shot target task without labeled samples. 
\vspace{-0.5cm}
\paragraph{Domain Adaptation:} The main focus of domain adaptation is to transfer domain knowledge from a data-rich domain to a domain with limited data \cite{luo2017label}\cite{csurka2017domain}. Learning domain-invariant features requires domain alignment. Such matching is done either by mid-level features of a CNN \cite{ghifary2016deep},  using an autoencoder \cite{ghifary2016deep}, by clustering \cite{sener2016learning}, or more recently, by using generative adversarial networks \cite{liu2016coupled}. In some recent efforts \cite{saito2017maximum}\cite{chen2018re}, source and target domain discrepancy is learned in an unsupervised manner. However, a considerable amount of labeled data from both domains is still unavoidable. 
In our methodology, we propose a generalizable framework that can learn models for a novel task from the knowledge of available tasks and their correlation with novel tasks.
\vspace{-0.5cm}
\paragraph{Meta-Learning:} Earlier efforts on meta-learning (with other objectives) assume that task parameters lie on a low-dimensional subspace \cite{argyriou2008convex}, share a common probabilistic prior \cite{lee2007learning}, etc. Unfortunately, these efforts are targeted only to achieve knowledge transfer among known tasks and tasks with limited data. Recent meta-learning approaches consider all task parameters as input signals to learn a meta manifold that helps few-shot learning \cite{naik1992meta}, \cite{thrun2012learning}, transfer learning \cite{redmon2016you} and domain adaptation \cite{ghifary2016deep}. A recent approach introduces learning a meta model in a model-agnostic manner \cite{finn2017model}\cite{kim2018bayesian} such that it can be applied to a variety of learning problems. Unfortunately, all these methods depend on the availability of a certain amount of labeled data in target domain to learn the transfer function, and cannot be scaled to novel tasks with no labeled data. Besides, the meta manifold learned by these methods are not explicit enough to extrapolate parameters of zero-shot tasks. Our method relaxes the need for ground truth by leveraging task correlation among known tasks and novel tasks. To the best of our knowledge, this is the first such work that involves regressing model parameters of novel tasks without using any ground truth information for the task.
\vspace{-0.5cm}
\paragraph{Learning with Weak Supervision:} Task correlation is used as a form of weak supervision in our methodology. Recent methods such as \cite{ratner2016data}\cite{varma2016socratic} proposed generative models that use a fixed number of user-defined weak supervision to programatically generate synthetic labels for data in near-constant time. Alfonseca \textit{et al.} \cite{alfonseca2012pattern} use heuristics for weak supervision to acccomplish hierachical topic modeling. Broadly, such weak supervision is harvested from knowledge bases, domain heuristics, ontologies, rules-of-thumb, educated guesses, decisions of weak classifiers or obtained using crowdsourcing. Structure learning \cite{bach2017learning} also exploits the use of distant supervision signals for generating labels. Such methods use factor graph to learn a high fidelity aggregation of crowd votes. Similar to this, \cite{pal2018adversarial} uses weak supervision signals inside the framework of a generative adversarial network. However, none of them operate in a zero-shot setting.
We also found related work zero-shot task generalization in the context of reinforcement learning (RL) \cite{oh2017zero}, or in lifelong learning \cite{isele2016using}. An agent is validated based on its performance on unseen instructions or a longer instructions. We find that the interpretation of task, as well as the primary objectives, are very different from our present course of study. 
\section{Methodology}
\label{sec_methodology}
The primary objective of our methodology is to learn a meta-learning algorithm that regresses nearly optimum parameters of a novel task for which no ground truth (data or labels) is available. To this end, our meta-learner seeks to learn from the model parameters of known tasks (with ground truth) to adapt to a novel zero-shot task. Formally, let us consider $K$ tasks to accomplish, i.e. $\mathcal{T} = {\tau_1, \cdots, \tau_K }$, each of whose model parameters lie on a meta-manifold $\mathcal{M}_\theta$ of task model parameters. 
We have ground-truth available for first $m$ tasks, i.e. \{$\tau_1, \cdots, \tau_m$\}, and we know their corresponding model parameters \{$(\theta_{\tau_i}): i = 1, \cdots, m$\} on $\mathcal{M}_\theta$. Complementarily, we have no knowledge of the ground truth for the zero-shot tasks \{$\tau_{(m+1)}, \cdots, \tau_K$\}. (We call the tasks \{$\tau_1, \cdots, \tau_m$\} as known tasks, and the rest \{$\tau_{(m+1)}, \cdots, \tau_K$\} as zero-shot tasks for convenience.)
Our aim is to build a meta-learning function $\mathcal{F(\cdot)}$ that can regress the unknown zero-shot model parameters \{$(\theta_{\tau_j}): j = (m+1), \cdots, K$\} from the knowledge of known model parameters (see Figure \ref{fig_2_final_figure_manifold} (b), i.e.: 
\vspace{-3pt}
\begin{equation}
    \label{eq_meta_1}
    \mathcal{F}(\theta_{\tau_1}, \cdots, \theta_{\tau_m}) = \theta_{\tau_j}, \quad j=m+1,\cdots,K
\end{equation}
\vspace{-3pt}
However, with no knowledge of relationships between the tasks, it may not be plausible to learn $\mathcal{F(\cdot)}$ as it can map to any point on $\mathcal{M}_\theta$. We hence introduce a task correlation matrix, $\Gamma$, where each entry $\gamma_{i,j} \in \Gamma$ captures the task correlation between two tasks $\tau_i, \tau_j \in \mathcal{T}$. 
Equation \ref{eq_meta_1} hence now becomes:
\vspace{-3pt}
\begin{equation}
    \label{eq_meta_2}
    \mathcal{F}(\theta_{\tau_1}, \cdots, \theta_{\tau_m}, \Gamma) = \theta_{\tau_j}, \quad j=m+1,\cdots,K
\end{equation}
\vspace{-3pt}
The function $\mathcal{F(.)}$ is itself parameterized by $W$. We design our objective function to compute an optimum value for $W$ as follows:
\vspace{-3pt}
\begin{equation}
    \label{eq_meta_3}
    \begin{split}
    & \min_{W} \sum_{i=1}^{m} ||\mathcal{F}((\theta_{\tau_1}, \gamma_{1,i}), \cdots, (\theta_{\tau_m}, \gamma_{m,i})); W) - \theta_{\tau_i}^{*}||^{2}
    \end{split}
\end{equation}
\vspace{-3pt}
Similar to \cite{zamir2018taskonomy}, without any loss of generality, we assume that all task parameters are learned as an autoencoder. Hence, our previously mentioned task parameters $\theta_{\tau_i}$ can be described in terms of an encoder, i.e. $\theta_{E_{\tau_i}}$, and a decoder, i.e. $\theta_{D_{\tau_i}}$. We observed that considering only encoder parameters $\theta_{E\tau_i}$ in Equation \ref{eq_meta_3} is sufficient to regress zero-shot encoders and decoders for tasks \{$\tau_{(m+1)}, \cdots, \tau_K$\}. Based on this observation, we rewrite our objective as (we show how our methodology works with other inputs in later sections of the paper):
\vspace{-3pt}
\begin{equation}
    \label{eq_meta_4}
    \begin{split}
    & \quad \min_{W} \sum_{i=1}^{m} ||\mathcal{F}\bigg((\theta_{E_{\tau_1}}, \gamma_{1,i}), \cdots, (\theta_{E_{\tau_1}}, \gamma_{m,i}); W\bigg)\\
    & \quad - (\theta_{E_{\tau_i}}^{*}, \theta_{D_{\tau_i}}^{*}) ||^{2}
    \end{split}
\end{equation}
\vspace{-3pt}
\noindent where $\theta_{E_{\tau_i}}^{*}$ and $\theta_{D_{\tau_i}}^{*}$ and the learned model parameters of a known task $\tau_i \in \mathcal{T}$. 
This alone is, however, insufficient. The model parameters thus obtained not only should minimize the above loss function on the meta-manifold $\mathcal{M}_{\theta}$, but should also have low loss on the original data manifold (ground truth of known tasks).

Let $\mathcal{D_{\theta_{D\tau_i}}(.)}$ denote the data decoder parametrized by $\theta_{D\tau_i}$, and $\mathcal{E_{\theta_{E\tau_i}}(.)}$ denote the data encoder parametrized by $\theta_{E\tau_i}$. We now add a \textit{data model consistency loss} to Equation \ref{eq_meta_4} to ensure that our regressed encoder and decoder parameters  perform well on both the meta-manifold network as well as the original data network:
\vspace{-3pt}
\begin{equation}
    \label{eq_meta_5}
    \begin{split}
    & \quad \min_{W} \sum_{i=1}^{m} ||\mathcal{F}\bigg((\theta_{E_{\tau_1}}, \gamma_{1,i}), \cdots, (\theta_{E_{\tau_1}}, \gamma_{m,i}); W\bigg)\\
    & \quad - (\theta_{E_{\tau_i}}^{*}, \theta_{D_{\tau_i}}^{*}) ||^{2}\\
    & \quad \quad + \lambda \sum_{\substack{x \in X_{\tau_i} \\ y \in \mathbf{y}_{\tau_i}}} \mathcal{L}{\quad \bigg(\mathcal{D}_{\tilde{\theta}_{D\tau_i}}(\mathcal{E}_{\tilde{\theta}_{E\tau_i}}(x)), y\bigg)}
    \end{split}
\end{equation}
\vspace{-3pt}
\noindent where $\mathcal{L}(\cdot)$ is an appropriate loss function (mean-squared error, cross-entropy or similar) defined for the task $\tau_i$.
\vspace{-0.5pt}
\paragraph{Network:} To accomplish the aforementioned objective in equation \ref{eq_meta_5}, we design $\mathcal{F}(.)$ as a network of $m$ branches, each with parameters \{$W_1, \cdots, W_m$\} respectively. These are not coupled in the initial layers but are later combined in a $W_{common}$ block that regresses encoder and decoder parameters. Dividing $\mathcal{F}(.)$ into two parts, $W_i$s and $W_{common}$, is driven by the intuition discussed in \cite{yosinski2014transferable}, that initial layers of $\mathcal{F(.)}$ transform the individual task model parameters into a suitable representation space, and later layers parametrized by $W_{common}$ capture the relationships between tasks and contribute to regressing the encoder and decoder parameters. For simplicity, we refer $W$ to mean \{$W_1, \cdots, W_m$\} and $W_{common}$. More specifics of the architecture of our model, \model, are discussed as part of our implementation details in Section \ref{sec_result}.
\vspace{-0.5pt}
\paragraph{Learning Task Correlation:} Our methodology admits any source of obtaining task correlation, including through other work such as \cite{zamir2018taskonomy}. In this work, we obtain the task correlation matrix, $\Gamma$, using crowdsourcing. Obtaining task relationships from wisdom-of-crowd (and subsequent vote aggregation) is fast, cheap, and allows several inputs such as rule-of-thumb, ontologies, domain expertise, etc. 
We obtain correlations for commonplace tasks used in our experiments from multiple human users. The obtained crowd votes are aggregated using the Dawid-Skene algorithm \cite{dawid1979maximum} to provide a high fidelity task relationship matrix, $\Gamma$.
\vspace{-0.5pt}
\paragraph{Input:} To train our meta network $\mathcal{F(.)}$, we need a batch of model parameters for each known task $\tau_1,\cdots,\tau_m$. This process is similar to the way a batch of data samples are used to train a standard data network. To obtain a batch of $p$ model parameters for each task, we closely follow the procedure described in \cite{Wang2017LearningTM}. This process is as follows. In order to obtain one model parameter set $\Theta^*_{\tau_i}$, for a known task $\tau_i$, we train a base learner (autoencoder), defined by $\mathcal{D}(\mathcal{E}(x; \theta_{E_{\tau_i}}); \theta_{D_{\tau_i}})$. This is achieved by optimizing the base learner on a subset (of size $l$) of data $\mathbf{x} \in X_{\tau_i}$ and corresponding labels $y \in \mathbf{y}_{\tau_i} $ with an appropriate loss function for the known task (mean-square error, cross-entropy or the like, based on the task). Hence, we learn one $\Theta^{*1}_{\tau_i} = \{\theta_{E\tau_i}^{*1}, \theta_{D\tau_i}^{*1}\}$. Similarly, $p$ subsets of labeled data are obtained using a sampling-with-replacement strategy from the dataset $( X_{\tau_i}, \mathbf{y}_{\tau_i})$ corresponding to $\tau_j$. Following this, we obtain a set of $p$ optimal model parameters (one for each of $p$ subsets sampled), i.e. $\Theta^{*}_{\tau_j}= {\Theta^{*1}_{\tau_j}, \cdots, \Theta^{*p}_{\tau_j}}$, for task $\tau_{j}$. A similar process is followed to obtain $p$ ``optimal'' model parameters for each known task $\{\Theta^*_{\tau_1}, \cdots, \Theta^*_{\tau_m}\}$. These model parameters (a total of $p \times m$ across all known tasks) serve as the input to our meta network $\mathcal{F(.)}$.

\vspace{-0.5pt}
\paragraph{Training:} The meta network $\mathcal{F(.)}$ is trained on the objective function in Eqn \ref{eq_meta_5} in two modes: a \textit{self mode} and a \textit{transfer mode} for each task. Given a known task $\tau_i$, training in \textit{self mode} implies updation of weights $W_i$ and $W_{common}$ alone. 
On the other hand, training in \textit{transfer mode} implies updation of weights $W_{\lnot i}$ (all $W_{j \neq i}, j = 1,\cdots,m$) and $W_{common}$ of $\mathcal{F(.)}$. \textit{Self mode} is similar to training a standard autoencoder, where $\mathcal{F(.)}$ leanrs to projects the model parameters $\theta_{\tau_j}$ near the given model parameter (learned from ground truth) $\theta^*_{\tau_j}$. In \textit{transfer mode}, a set of model parameters of tasks (other than $\tau_j$) attempt to map the position of learned $\theta_{\tau_j}$, near the given model parameter $\theta_{\tau_j}$ on the meta manifold. We note that the transfer mode is essential in being able to regress model parameters of a task, given model parameters of other tasks. At inference time (for zero-shot task transfer), $\mathcal{F(.)}$ operates in transfer mode. 
\begin{figure}
 \centering 
 \includegraphics[width=8cm]{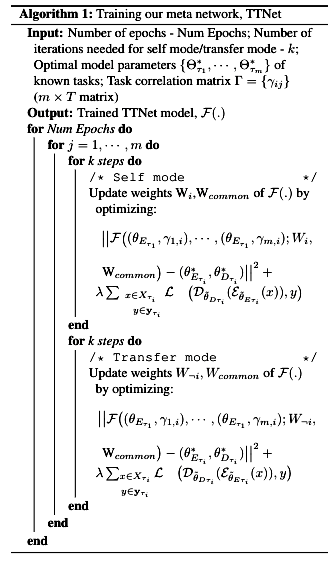}
 \label{fig_zero_shot_results}
\end{figure}
\vspace{-0.5pt}
\paragraph{Regressing Zero-Shot Task Parameters:} Once we learn the optimal parameters $W^*$ for $\mathcal{F(.)}$ using Algorithm \ref{alg_TTNet_training}, we use this to regress zero-shot task parameters, i.e.  $\mathcal{F}_{W^*}\big((\theta_{E\tau_1}, \gamma_{1,j}), \cdots, (\theta_{E\tau_m}, \gamma_{m,j})\big)$ for all $j = (m+1), \cdots, T$. (We note that the implementation of Algorithm 1 was found to be independent of the ordering of the tasks, $\tau_1, \cdots, \tau_m$.)
\section{Results}
\label{sec_result}
\vspace{-3pt}
To evaluate our proposed framework, we consider the vision tasks defined in \cite{zamir2018taskonomy}. (Whether this is an exhaustive list of vision tasks is arguable, but they are sufficient to support our proof of concept.) In this section, we consider four of the tasks as unknown or zero-shot: surface normal, depth estimation, room layout, and camera-pose estimation. We have curated this list based on the data acquisition complexity and the complexity associated with the learning process using a deep network. Surface normal, depth estimation and room layout estimation tasks are monocular tasks but involve expensive sensors to get labeled data points. Camera pose estimation requires multiple images (two or more) to infer six degrees-of-freedom and is generally considered a difficult task. We have four different \model s to accomplish them; (1) $\text{\model}_{6}$ considers 6 vision tasks as known tasks; (2) $\text{\model}_{10}$ considers 10 vision tasks as known tasks; and (3) $\text{\model}_{20}$ considers 20 vision tasks as known tasks. In addition, we have another model $\text{\model}_{LS}$ (20 known tasks) in which, the regressed parameters are finetuned on a small amount, (20\%), of data for the zero-shot tasks. (This provides 
low supervision and hence, the name $\text{\model}_{LS}$.) Studies on other sets of tasks as zero-shot tasks are presented in Section \ref{sec_discussion}. We also performed an ablation study on permuting the source tasks differently, which is presented in the supplementary section due to space constraints. 
\vspace{-3pt}
\subsection{Dataset}
\label{sub_sec_dataset}
\vspace{-4pt}
We evaluated \model on the Taskonomy dataset \cite{zamir2018taskonomy}, a publicly available dataset comprised of more than 150K RGB data samples of indoor scenes. It provides the ground truths of 26 tasks given the same RGB images, which is the main reason for considering this dataset. We considered 120K images for training, 16K images for validation, and, 17K images for testing. 
\subsection{Implementation Details}
\label{sub_implementation}
\vspace{-4pt}
\paragraph{Network Architecture:} Following Section \ref{sec_methodology}, each data network is considered an autoencoder, and closely follows the model architecture of \cite{zamir2018taskonomy}. The encoder is a fully convolutional ResNet 50 model without pooling, and the decoder comprises of 15 fully convolutional layers for all
pixel-to-pixel tasks, e.g. normal estimation, and for low dimensional
tasks, e.g. vanishing points, it consists of 2-3 FC layers. To make input samples for \model, we created 5000 samples of the model parameters for each task, each of which is obtained by training the model on 1k data points sampled (with replacement) from the Taskonomy dataset. These data networks were trained with mini-batch Stochastic Gradient Descent (SGD) using a batch size of 32, learning rate of 0.001, momentum factor of 0.5 and Adam as an optimizer.
\paragraph{\model:} \model's architecture closely follows the ``classification'' network of \cite{finn2017model}. We show our network 
is shown in Figure \ref{fig_2_final_figure_manifold} (b). The \model initially has $m$ branches, where $m$ depends on the model under consideration ($\text{\model}_m: m \in \{6, 10, 20\}$). Each of the $m$ branches is comprised of 15 fully convolutional (FCONV) layers followed by 14 fully connected layers. The $m$ branches are then merged to form a common layer comprised of 15 FCONV layers. We trained the complete model with mini-batch SGD using a batch size of 32, learning rate of 0.0001, momentum factor of 0.5 and Adam as an optimizer.

\paragraph{Task correlation:} Crowds are asked to response for each pair of tasks (known and zero) on a scale of $+2$ (strong correlation) to $-1$ (no correlation), while $+3$ is reserved to denote self relation. We then aggregated crowd votes using Dawid-skene algorithm which is based on the principle of Expectation-Maximization (EM). More details of the Dawid-skene methodology and vote aggregation are deferred to the supplementary section.
\subsection{Comparison with State-of-the-Art Models}
\vspace{-3pt}
\begin{figure*}
 \centering 
 \includegraphics[width=17cm]{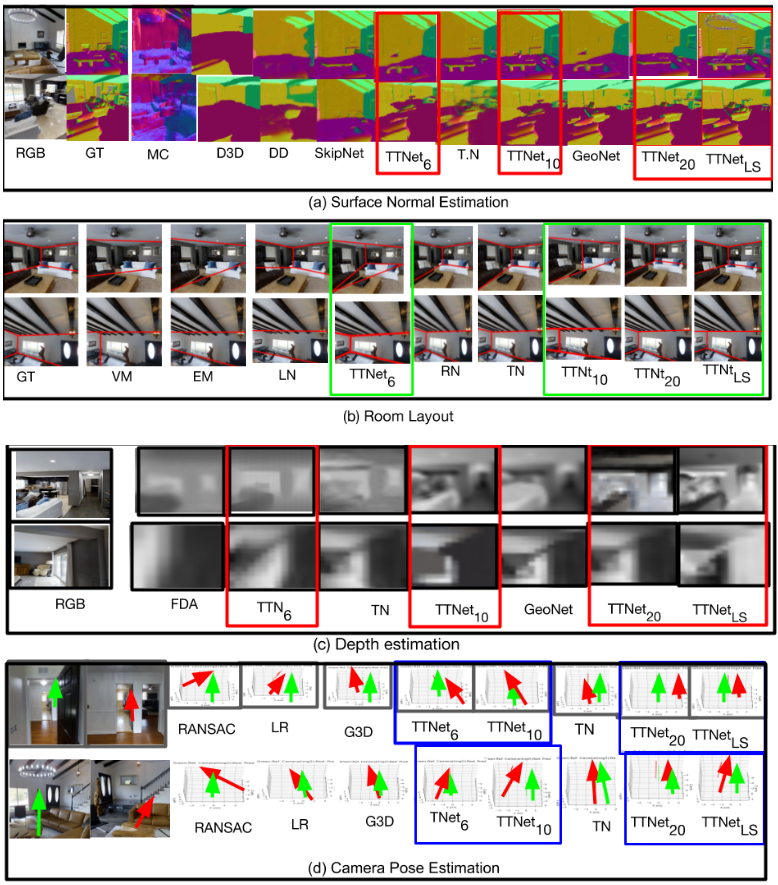}
 \caption{\textbf{Qualitative comparison (Best viewed in color):} \model models compared against other state-of-the-art models, see Section \ref{subsec_qualitative_results} for details.  (a) \textit{\textbf{Surface Normal Estimation:}} Red boxes indicate results of our \model models; (b) \textit{\textbf{Room Layout:}} Red edges indicate the predicted room edges; green boxes indicate our \model model results; (c) \textit{\textbf{Depth Estimation:}} Red bounding boxes show our results; (d) \textit{\textbf{Camera Pose Esimation:}} First image is the reference frame of the camera, i.e. green arrow. The second image, with red arrow, is taken after a geometric translation w.r.t first image. Blue rectangles show our results.}
 \label{fig_zero_shot_results}
\end{figure*}
We show both qualitative and quantitative results for our \model, trained using the aforementioned methodology, on each of the four identified zero-shot tasks against state-of-the-art models for each respective task below. We note that the same \model is validated against all tasks.
\subsubsection{Qualitative Results}
\label{subsec_qualitative_results}
\vspace{-3pt}
\paragraph{Surface Normal Estimation:} For this task, our \model is compared against the following state-of-the-art models: Multi-scale CNN (MC) \cite{eigen2015predicting}, Deep3D (D3D) \cite{wang2015designing}, Deep Network for surface normal estimation (DD) \cite{wang2015designing}, SkipNet \cite{bansal2016marr}, GeoNet \cite{qi2018geonet} and Taskonomy (TN) \cite{zamir2018taskonomy}. The results are shown in Figure \ref{fig_zero_shot_results}(a), where the red boxes correspond to our models trained under different settings (as described at the beginning of Section \ref{sec_result}. It is evident from the result that $\text{\model}_{6}$ gives visual results similar to \cite{zamir2018taskonomy}. As we increase the number of source tasks, our \model shows improved results. $\text{\model}_{WS}$ captures finer details (see edges of chandelier) which is not visible in any other result.  
\vspace{-3pt}
\paragraph{Room Layout Estimation:} We followed the definition of layout types in \cite{lee2017roomnet}, and our \model's results are compared against following camera pose methods: Volumetric \cite{gupta2010estimating}, Edge Map \cite{zhang2017learning}, LayoutNet \cite{zou2018layoutnet}, RoomNet \cite{lee2017roomnet}, and Taskonomy \cite{zamir2018taskonomy}. The green boxes in Figure \ref{fig_zero_shot_results}(b) indicate \model results; the red edges indicate the predicted room edges. Each model infers room corner points and joins them with straight lines. We report two complex cases in Figure \ref{fig_zero_shot_results} (b): (1) lot of occlusions, and (2) multiple edges such as roof-top, door, etc. 
\vspace{-3pt}
\paragraph{Depth Estimation:} Depth is computed from a single image. We compared our \model against: FDA \cite{lee2018single}, Taskonomy \cite{zamir2018taskonomy}, and GeoNet \cite{qi2018geonet}. The red bounding boxes show our result. It can be observed from Figure \ref{fig_zero_shot_results}(c) that $\text{\model}_{10}$ outperforms \cite{zamir2018taskonomy}; and $\text{\model}_{20}$ and $\text{\model}_{WS}$ outperform all other methods studied.
\vspace{-3pt}
\paragraph{Camera Pose Estimation (fixed):} Camera pose estimation requires two images captured from two different geometric points of view of the same scene. A fixed camera pose estimation predicts any five of the 6-degrees of freedom: yaw, pitch, roll and x,y,z translation. In Figure \ref{fig_zero_shot_results}(d), we show two different geometric camera angle translations: (1) perspective, and (2) translation in y and z coordinate. First image is the reference frame of the camera, i.e. green arrow. The second image, i.e. the red arrow, is taken after a geometric translation w.r.t the first image. We compared our model against: RANSAC \cite{derpanis2010overview}, Latent RANSAC \cite{korman2018latent}, Generic3D pose \cite{zamir2016generic} and Taskonomy \cite{zamir2018taskonomy}. Once again, $\text{\model}_{20}$ and $\text{\model}_{WS}$ outperform all other methods studied.
\vspace{-3pt}
\subsubsection{Quantitative Results}
\label{sub_quality}
\vspace{-3pt}
\paragraph{Surface Normal Estimation:} We evaluated our method based on the evaluation criteria described in \cite{qi2018geonet}, \cite{bansal2016marr}. The results are presented in Table \ref{table_surface_normal}. 
Our $\text{\model}_{6}$ is comparable to state-of-the-art Taskonomy \cite{zamir2018taskonomy} and GeoNet \cite{qi2018geonet}. Our $\text{\model}_{10}$, $\text{\model}_{20}$, and $\text{\model}_{WS}$ outperforms all state-of-the-art models.

\begin{table}
\resizebox{0.9\columnwidth}{!}{%
\begin{tabular}{|p{1.3cm}|p{.7cm}|p{.7cm}|p{.7cm}|p{.7cm}|p{.7cm}|p{.7cm}|}
            \hline \hline
            Method&Mean ($\downarrow$)&Medn ($\downarrow$)&RMSE ($\downarrow$)& 11.25 ($\uparrow$)& 22.5 ($\uparrow$)& 30 ($\uparrow$)  \\ \hline \hline
            MC\cite{eigen2015predicting} & 30.30 & 35.30 & - & 30.29 & 57.17 & 68.29\\ \hline
            D3D \cite{wang2015designing}& 25.71 & 20.81 & 31.01 & 38.12 & 59.18 & 67.21\\ \hline
            DD\cite{wang2015designing} & 21.10 & 15.61 & - & 44.39 & 64.48 & 66.21\\ \hline
            SkipNet \cite{bansal2016marr} &20.21 & 12.19 & 28.20 & 47.90 & 70.00 & 78.23\\ \hline
            TN\cite{zamir2018taskonomy}& 19.90 & 11.93 & 23.13 & 48.03 & 70.02 & 78.88\\ \hline
            TTNet$_{6}$& 19.22 & 12.01 & 26.13 & 48.02 & 71.11 & 78.29\\ \hline
            GeoNet \cite{qi2018geonet}& 19.00 & 11.80 & 26.90 & 48.04 & 72.27 & 79.68\\ \hline
            TTNet$_{10}$& 19.81 & 11.09 & 22.37 & 48.83 & 71.61 & 79.00\\ \hline
            TTNet$_{20}$& 19.27 & 11.91 & 26.44 & 48.81 & 71.97 & 79.72\\ \hline
            TTNet$_{LS}$& \textbf{15.10} & \textbf{9.29} & \textbf{24.31} & \textbf{56.11} & \textbf{75.19} & \textbf{84.71} \\ \hline \hline
        \end{tabular}%
        }
    \caption{\textbf{Surface Normal Estimation.} Mean, median and RMSE refer to the difference between the model's predicted surface normal and ground truth surface normal (a lower value is better). Other 3 are the number of pixels within degree 11.25, 22.5 and 30 thresholds within ground truth's predicted pixels (a higher number is better). $-$ indicates those values cannot be obtained by the corresponding method.}
\label{table_surface_normal}
\end{table}
\vspace{-3pt}
\paragraph{Room Layout Estimation:} We use two standard evaluation criteria: (1) \textit{Keypoint error:} a global measurement avaraged on Euclidean distance between model's predicted keypoint and the ground truth; and (2) \textit{Pixel error:} a local measurement that estimates pixelwise error between the predicted surface labels and ground truth labels. Table \ref{table_room_layout} presents the results. A lower number corresponding to our \model models indicate good performance.
\begin{table}[h]
    \resizebox{0.97\columnwidth}{!}{%
        \begin{tabular}{|p{1cm}|p{0.7cm}|p{0.7cm}|p{0.7cm}|p{1cm}|p{0.7cm}|p{0.7cm}|p{0.9cm}|p{0.9cm}|p{0.9cm}|}
            \hline \hline
            Methd& VM \cite{gupta2010estimating}& EM \cite{zhang2017learning}& LN \cite{zou2018layoutnet}&TTNet$_{6}$&RN \cite{lee2017roomnet}&TN \cite{zamir2018taskonomy}&TTNt$_{10}$&TTNt$_{20}$&TTNt$_{LS}$\\ \hline \hline
            Keypt.&15.48&11.2&7.64&7.51&6.30&6.22&6.00&5.82&\textbf{5.52}\\ \hline
            Pixel&24.33&16.71&10.63&8.10&8.00&8.00&7.72&7.10&\textbf{6.81}\\ \hline \hline
        \end{tabular}
    }
    \caption{\textbf{Room Layout.} Both $\text{\model}_{20}$ and $\text{\model}_{LS}$ outperformed state-of-the-art models on keypoint and pixel error.}
\label{table_room_layout}
\end{table}
\vspace{-3pt}
\paragraph{Depth Estimation:} We followed the evaluation criteria for depth estimation as in \cite{lee2018single}, where the metrics are: RMSE (lin) = $\frac{1}{N} (\sum_{X} (d_{X} - d^*_{X})^2)^{\frac{1}{2}}$; RMSE(log) =  $\frac{1}{N}(\sum_{X} (\log d_{X} - \log d^*_{X})^2)^{\frac{1}{2}}$; Absolute relative distance = $\frac{1}{N}\sum_{X}\frac{|d_{X} - d^*_{X}|}{d_{X}}$; Squared absolute relative distance = $\frac{1}{N}\sum_{X} \big(\frac{|d_{X} - d^*_{X}|}{d_{X}}\big)^2$. Here, $d^{*}_{X}$ is ground truth depth, $d_{X}$ is estimated depth, and $N$ is the total number of pixels in all images in the test set.
\begin{table}[h]
    \centering
    \small
        \begin{tabular}
            {|p{1.3cm}|c|c|c|c|c|}
            \hline \hline
            Method & RMSE(lin)& RMSE(log) & ARD & SRD\\ \hline \hline
            FDA \cite{lee2018single} & 0.877 &0.283 &0.214 &0.204\\ \hline
            TTN$_6$ &0.745& 0.262& 0.220& 0.210\\ \hline
            TN \cite{zamir2018taskonomy}&0.591& 0.231& 0.242& 0.206\\ \hline
            TTNt$_{10}$ & 0.575 &0.172 &0.236 &0.179\\ \hline
            Geonet\cite{qi2018geonet}& 0.591&0.205 &0.149 &0.118\\ \hline
            TTNet$_{20}$ & 0.597 &0.204 &0.140 &0.106\\ \hline
            TTNet$_{LS}$& \textbf{0.572} &\textbf{0.193} &\textbf{0.139} &\textbf{0.096}\\ \hline \hline
        \end{tabular}
    \caption{\textbf{Depth estimation:}  $\text{\model}_{20}$ and $\text{\model}_{WS}$ outperform all other methods studied.}
\label{table_inpainting}
\end{table}
\vspace{-3pt}
\paragraph{Camera Pose Estimation (fixed):} We adopted the \textit{win rate} (\%) evaluation criteria \cite{zamir2018taskonomy} that counts the proportion of images for which a baseline is outperformed. Table \ref{table_camera_pose} shows the win rate of \model models on angular error with respect to state-of-the-art models: RANSAC \cite{yosinski2014transferable}, LRANSAC \cite{korman2018latent}, G3D and Taskonomy \cite{zamir2018taskonomy}. The results show the promising performance of \model.
\begin{table}[h]
    \footnotesize
        \begin{tabular}
            {|c|c|c|c|c|}
            \hline \hline
           \small{Method}&\small{RANSAC}\cite{yosinski2014transferable}& LR\cite{korman2018latent}& G3D\cite{zamir2016generic} & TN\cite{zamir2018taskonomy}\\
            \hline \hline
            TTNet$_{6}$& 88\%&81\%&72\%&64\%\\ \hline
            TTNet$_{10}$& 90\%&82\%&79\%&82\%\\ \hline
            TTNet$_{20}$& 90\%&82\%&92\%&80\%\\ \hline
            TTNet$_{LS}$& \textbf{96\%}&\textbf{88\%}&\textbf{96\%}&\textbf{87\%}\\ \hline \hline
        \end{tabular}
    \caption{\textbf{Camera Pose Estimation (fixed)}. We have considered \textit{win rate} (\%) on angular error. Columns are state-of-the-art methods and rows are our four \model models. }
\label{table_camera_pose}
\end{table}
\begin{figure}
 \centering 
 \includegraphics[width=9cm]{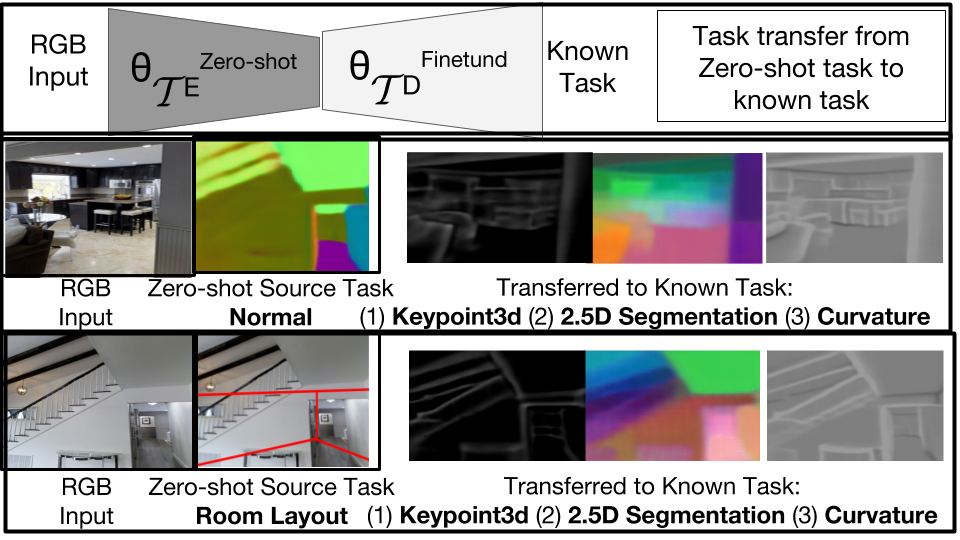}
 \caption{\textbf{Zero-shot task to known task transfer.} We consider the zero-shot tasks: \textit{surface normal estimation} and \textit{room layout estimation}, and transfer to models for Keypoint 3D, 2.5D segmentation and curvature estimation. }
 \label{fig_zero_shot_to_known_task_transfer}
\end{figure}
\section{Discussion and Analysis}
\label{sec_discussion}
\begin{figure*}[h]
    \centering
    \includegraphics[width=17cm, height=5.5cm]{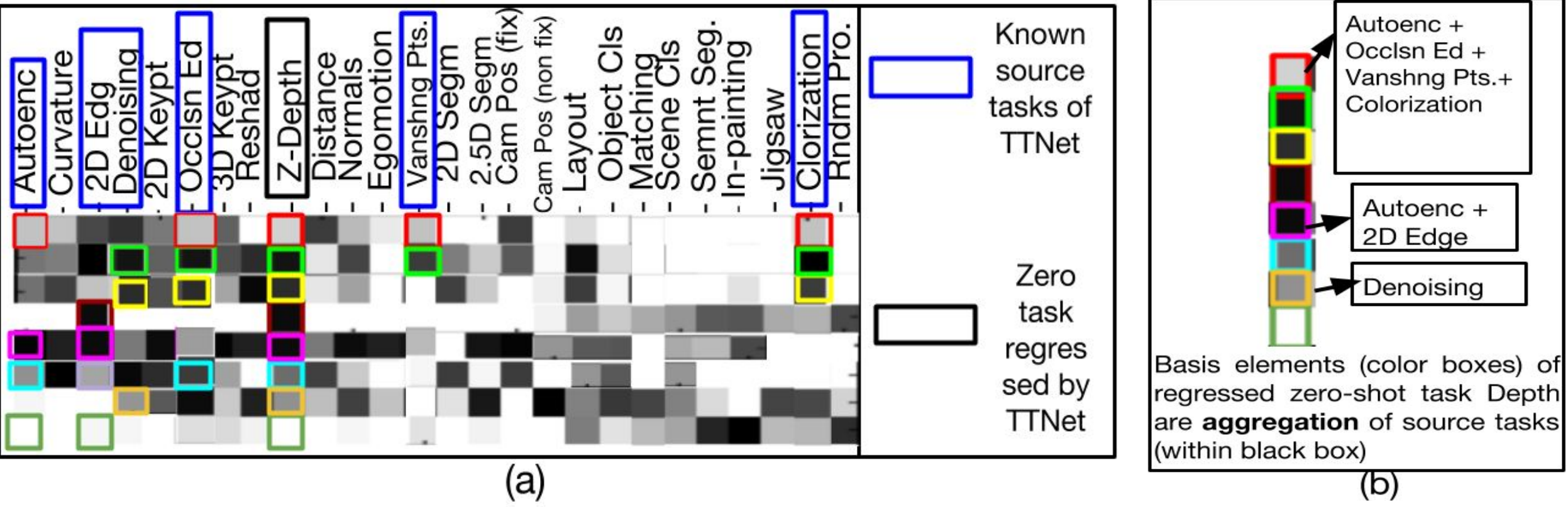}
    \caption{\textbf{Finding the basis of tasks:} Latent task basis are estimated following the GO-MTL approach \cite{kumar2012learning}. \textbf{(a)} \textit{Most important source} has the highest number of basis elements with similar values as zero-shot task. Example: source task ``Autoencoding'' (col $1$) is important for zero-shot task ``Z-Depth'' (col $9$) as they share 4 such basis elements. \textbf{(b)} When tasks are related (which is the setting in our work), learning from similar tasks can by itself provide good performance. Example: basis vector of zero-shot task ``Depth'' is composed of latent elements several source tasks.}
    \label{fig:basis}
\end{figure*}

\vspace{-7pt}
\paragraph{Significance Analysis of Source Tasks:} An interesting question on our approach is: how do we quantify the contribution of each individual source task towards regressing parameter of target task? In other word, which source task plays the most important role to regress the zero-shot task parameter. Figure \ref{fig:basis} quantifies this by considering latent task basis to estimate this. We followed GO-MTL approach \cite{kumar2012learning} to compute the task basis. Optimal model parameters of known tasks are mapped to a low-dim vector space $R$ using an autoencoder, before applying GO-MTL. 

Formally speaking, optimal model parameters of each known task are mapped to a low-dimensional space $R$, i.e. $\mathcal{S}: \Theta_{\tau_i} \rightarrow R_i$. $S(.)$ using an autoencoder trained on model parameters of known tasks $\{\Theta^*_{\tau_1}, \cdots, \Theta^*_{\tau_m}\}$, i.e. $\min_{J} \sum_{i=1}^{m} ||\mathcal{S}(\Theta_{\tau_i}; J) - \Theta_{\tau_i}^{*}||^{2} + \lambda \sum_{\substack{(x,y) \in (X_{\tau_i},\mathbf{y}_{\tau_i})}} \mathcal{L}{ \big(\mathcal{D}_{\tilde{\Theta}_{D\tau_i}}(\mathcal{E}_{\tilde{\Theta}_{E\tau_i}}(x)), y\big)}$ (similar to Eqn 5). $\mathcal{S}(.)$ infers latent representation $R_{zero}$ for regressed model parameter of zero-shot task $\Theta_{\tau_{zero}}$. We used ResNet-18 both for encoder-decoder, dimension of $R$ as 100, and the dimension of task basis as 8. We can then have task matrix \textbf{W}$_{100X26}$ = \textbf{L}$_{100X8}$\textbf{S}$_{8X26}$, comprised of all $R_i$ and $R_{zero}$. In Figure \ref{fig:basis}, boxes of same color denote similar-valued weights of task basis vectors. \textit{Most important source} has the highest number of basis elements with similar values as zero-shot task. In Figure \ref{fig:basis} (a) below, source task ``Autoencoding'' (col $1$) is important for zero-shot task ``Z-Depth'' (col $9$) as they share 4 such basis elements.
\paragraph{Why Zero-shot Task Parameters Performs Better than Supervised Training?} It is evident from our qualitative and quantitative study that regressed zero-shot parameters out-performs results from supervised learning. When tasks are related (which is the setting in our work), learning from similar tasks can by itself provide good performance. From Figure \ref{fig:basis}, we can see that, the basis vector of zero-shot task ``Depth'' is composed of latent elements several source tasks. 
E.g. in Figure \ref{fig:basis} (b) above, learning of 1$^{st}$ element (red box) of zero-shot task ``Z-depth'' is supported by 4 related source tasks.
\paragraph{Zero-shot to Known Task Transfer:} Are our regressed model parameters for zero-shot tasks capable of transferring to a known task? To study this, we consider the autoencoder-decoder parameters for a zero-shot task, and finetune the decoder to a target known task, following the procedure in \cite{zamir2018taskonomy} (encoder remains the same as of zero-shot task). Figure \ref{fig_zero_shot_to_known_task_transfer} shows the qualitative results, which are promising. We also compared our \model against \cite{zamir2018taskonomy} quantitatively by studying the \textit{win rate} (\%) of the two methods against other state-of-the-art methods: Wang \textit{et al.} \cite{Wang2017LearningTM}, G3D \cite{zamir2016generic}, and full supervision. Owing to space constraints, these results are presented in the supplementary section.
\vspace{-5pt}
\paragraph{Choice of Zero-shot Tasks:} In order to study the generalizability of our method, we conducted experiments with a different set of zero tasks than those considered in Section \ref{sec_result}.
\begin{figure}
 \centering 
 \includegraphics[width=8cm]{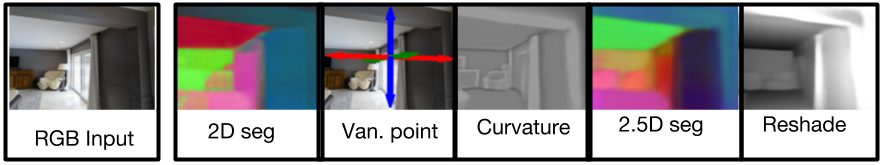}
 \caption{\textbf{Different Choice of Zero-Shot Tasks.} Results of $\text{\model}_6$ on different set of zero shot tasks: 2D segmentation, Vanishing point estimation, Curvature estimation, 2.5D segmentation and reshading.}
 \label{fig_zero_shot_rotation}
\end{figure}
Figure \ref{fig_zero_shot_rotation} shows promising results for our weakest model, $\text{\model}_{6}$, on other tasks as zero shot tasks. More results of our other models $\text{\model}_{10}$, $\text{\model}_{20}$, $\text{\model}_{WS}$ are included in the supplementary section.
\begin{figure}
 \centering 
 \includegraphics[width=8cm]{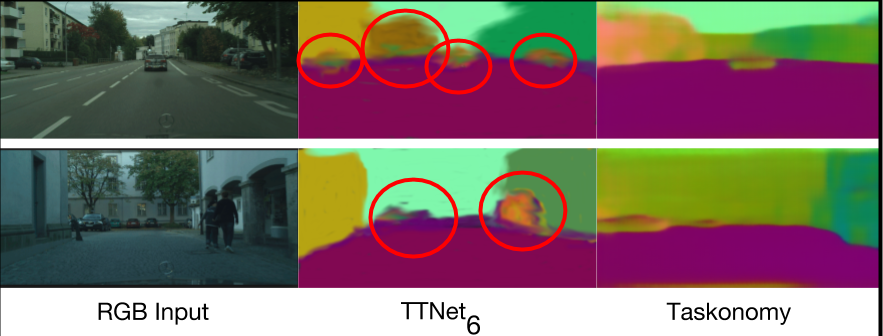}
 \caption{\textbf{Surface normal estimation on Cityscapes.} Red circles highlight details (car, tree, human) captured by our model, which is missed by Taskonomy}
 \label{fig_city_snf}
\end{figure}
\vspace{-6pt}
\paragraph{Performance on Other Datasets:} To further study the generalizability of our models, we finetuned \model on the Cityscapes dataset \cite{Cordts2016Cityscapes}, and the surface normal results are reported in Figure \ref{fig_city_snf}, with comparison to \cite{zamir2018taskonomy}. Our model captures more detail.
\paragraph{Object detection on COCO-Stuff dataset:} TTNet$_{6}$ is finetuned on the COCO-stuff dataset to do object detection on COCO-stuff dataset. To facilitate the object detection, we considered object classification as source task instead of colorization. TTNet$_{6}$ performs fairly well.
\begin{figure}[h]
    \centering
    \includegraphics[height=0.1\textwidth, width=0.4\textwidth]{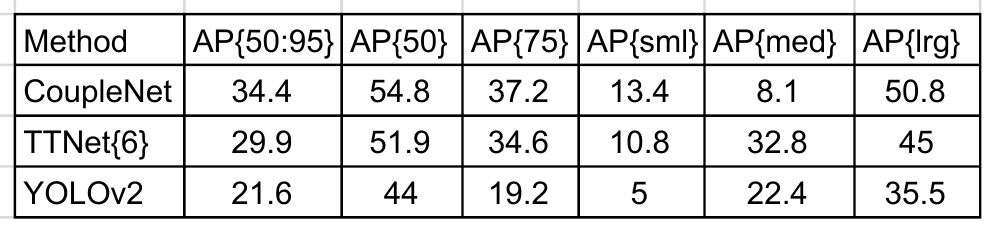}
    \caption{\textbf{Object Detection using TTNet$_{6}$}: TTNet$_{6}$ is finetuned on the COCO-stuff dataset to do object detection on COCO-stuff dataset.}
    \label{object_detection}
\end{figure}
\begin{table}[h]
\footnotesize
\begin{tabular}{|p{1.3cm}|p{0.5cm}|p{0.5cm}|p{0.5cm}|p{0.5cm}|p{0.5cm}|p{0.5cm}|p{0.7cm}|}
\hline
Model&TT$_4$&TT$_6$&TT$_8$&TT$_{10}$&TT$_{15}$&TT$_{20}$&TT$_{LS}$\\ \hline \hline
Wang\cite{Wang2017LearningTM}&81&84&84&88&88&91&97\\ \hline
Zamir\cite{zamir2016generic}&73&75&81&82&86&87&90\\ \hline
TN\cite{zamir2018taskonomy}&62&65&84&85&84&89&94\\ \hline \hline
\end{tabular}
\caption{\textit{Win rate} (\%) of surface normal estimation of \model models with varying num of known tasks against: \cite{Wang2017LearningTM}, \cite{zamir2016generic}, and \cite{zamir2018taskonomy}.}
\label{tab_optimal_tasks}
\end{table}
\vspace{-6pt}
\paragraph{Optimal Number of Known Tasks:} In this work, we have reported results of \model with 6, 10 and 20 known tasks. We studied the question - how many tasks are sufficient to adapt to zero-shot tasks in the considered setting, and the results are reported in Table \ref{tab_optimal_tasks}. Expectedly, a higher number of known tasks provided improved performance. A direction of our future work includes the study of the impact of negatively correlated tasks on zero-shot task transfer.

We also conducted experiments on using our methodology by using the task correlations obtained from the results of \cite{zamir2018taskonomy} directly. We present these, as well as other results, including the evolution of our \model model over the epochs of training, in the supplementary section.
\section{Conclusion}
\vspace{-4pt}
In summary, we present a meta-learning algorithm to regress model parameters of a novel task for which no ground truth is available (\textit{zero-shot task}). We evaluated our learned model on the Taskonomy \cite{zamir2018taskonomy} dataset, with four zero-shot tasks: surface normal estimation, room layout estimation, depth estimation and camera pose estimation. We conducted extensive experiments to study the usefulness of zero-shot task transfer, as well as showed how the proposed \model can also be used in transfer learning. Our future work will involve closer analysis of the implications of obtaining task correlation from various sources, and the corresponding results for zero-shot task transfer. In particular, negative transfer in task space is a particularly interesting direction of future work.
{\small
\bibliographystyle{ieee}
\bibliography{egpaper_for_review}
}
\clearpage
\begin{center}
\large{\textbf{Supplementary Section}}
\end{center}
In this section, we include more details on obtaining the task correlation matrix, $Gamma$, described in Section \ref{sec_methodology}, ablation studies with varying source tasks, as well as additional results and comparisons, which could not be included in the main paper due to space constraints.
\section{More on Task Correlation}
\paragraph{Dawid-Skene method:} As mentioned in Section \ref{sec_methodology}, we used the well-known Dawid-Skene (DS) method \cite{dawid1979maximum}\cite{Zhu2015OnlineC} to aggregate votes from human users to compute the task correlation matrix $\Gamma$. We now describe the DS method.

We assume a total of $M$ annotators providing labels for $N$ items, where each label belongs to one of $K$ classes. DS associates each annotator $m \in M$ with a $K \times K$ confusion matrix $\Theta^{m}$ to measure an annotator's performance. The final label is a weighted sum of annotator's decisions based on their confusion matrices, i.e. $\Theta^{m}$s where \{$m = 1, \cdots, M$\}. Each entry of the confusion matrix $\theta_{lg} \in \Theta^{m}$ is the probability of predicting class $g$ when the true class is $l$. A true label of an item $n \in N$ is $y_n$ and the vector $\textbf{y}$ denotes true label for all items, i.e. $\textbf{y}$ = \{$y_1, \cdots, y_N$\}. Let's denote $\xi_{m,n}$ as annotator $m$'s label for item $n$, i.e. if the annotator labeled the item as $k \in K$, we will write $\xi_{m,n}$ = $k$. Let matrix $\Xi$ $(\xi_{m,n} \in \Xi)$ denote all labels for all items given by all annotators. The DS method computes the annotators' error tensor $\mathcal{C}$, where each entry $c_{mlg} \in \mathcal{C}$ denotes the probability of annotator $m$ giving label $l$ as label $g$ for item $n$. The joint likelihood $\mathcal{L}(.)$ of true labels and observed labels $\Xi$ can hence be written as:
\begin{equation}
    \label{eq_dawid_1}
    \mathcal{L}(\mathcal{C};\textbf{y},\Xi) = \prod_{j=1}^{N}\prod_{m=1}^{M}\prod_{g=1}^{K} \big(c_{my_{j}g}\big)^{\textbf{1}_{(\xi_{m,j}=k)}}
\end{equation}
Maximizing the above likelihood provides us a mechanism to aggregate the votes from the annotators. To this end, we find the maximum likelihood estimate using Expectation Maximization (EM) for the marginal log likelihood below:
\begin{equation}
    \label{eq_dawid_2}
    l(\mathcal{C}) = \log( \sum\mathcal{L}(\mathcal{C};\textbf{y},\Xi))
\end{equation}
The E step is given by:
\begin{equation}
\centering
    \label{eq_dawid_3}
    \begin{split}
        & \quad \quad \mathbb{E}[\log \mathcal{L}(\mathcal{C};\textbf{y},\Xi)] = \\
        & \prod_{j=1}^{N} p(y_{j} = l |\mathcal{C},\Xi)\log\prod_{m=1}^{M}\prod_{g=1}^{K} \big(c_{mlg}\big)^{\textbf{1}_{(\xi_{m,j}=k)}}
    \end{split}
\end{equation}
The M step subsequently computes the $\mathcal{C}$ estimate that maximizes the log likelihood:
\begin{equation}
    \label{eq_dawid_4}
    \hat{c}_{mlg} = \frac{\prod_{j=1}^{N} p(y_{j} = l |\mathcal{C},\Xi)\textbf{1}_{(\xi_{m,j}=k)}}{\prod_{k^{'}=1}^{K}\Big(\prod_{j=1}^{N} p(y_{j} = l |\mathcal{C},\Xi)\textbf{1}_{(\xi_{m,j}=k^{'})}\Big)}
\end{equation}
\begin{equation}
    \label{eq_dawid_5}
    \hat{p}(y_j) = \frac{\prod_{j=1}^{N} \textbf{1}_{(\xi_{m,j}=k)}}{N}
\end{equation}
\begin{figure}
 \centering 
 \includegraphics[width=9cm]{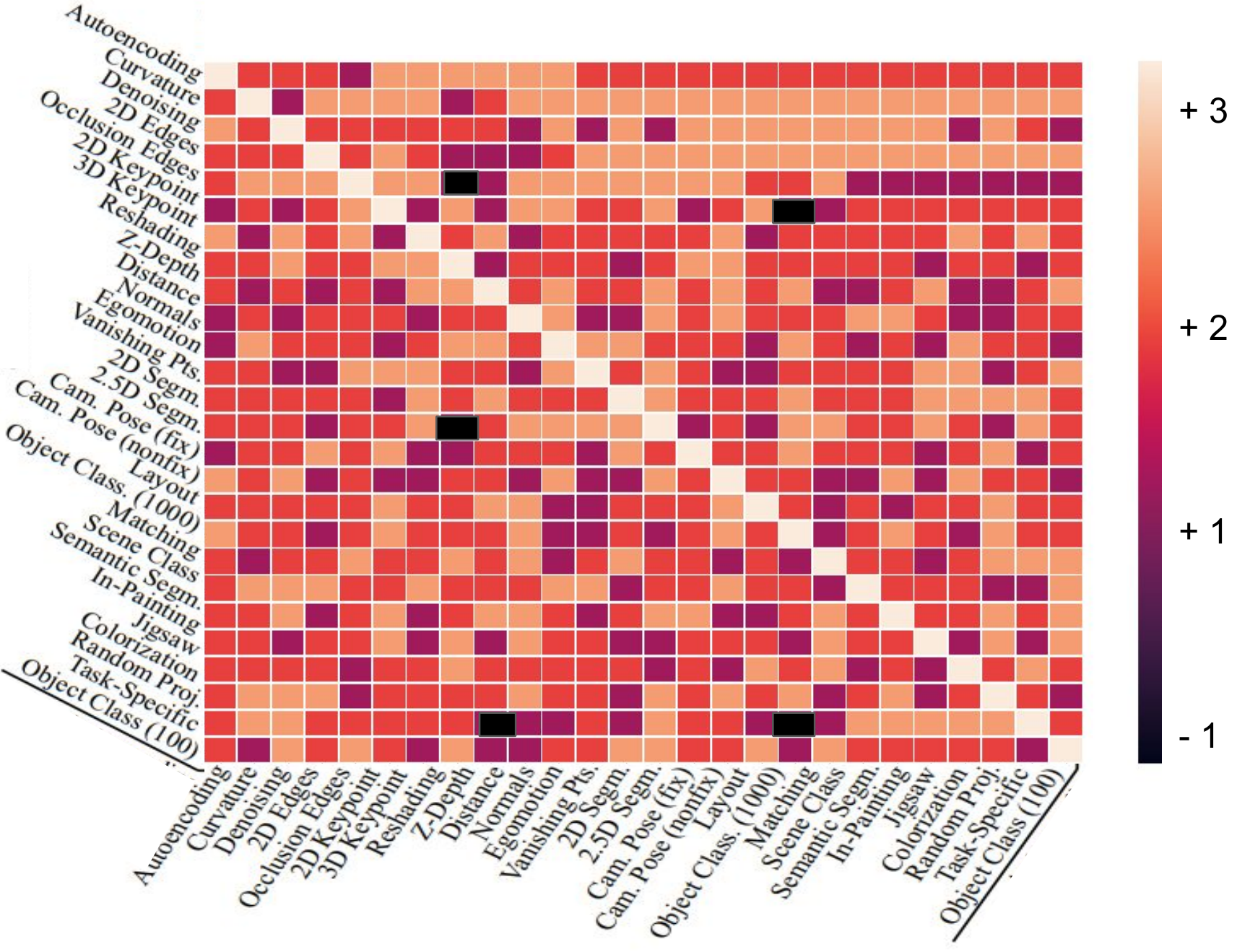}
 \caption{\textbf{Task correlation matrix.} We get the task correlation matrix $\Gamma$ after receiving votes from $30$ annotators. Annotators are asked to give task correlation label on a scale of \{$+3, +2, +1, 0, -1$\}. On that scale, a $+3$ denotes \textit{self relation}, $+2$ describes \textit{strong relation}, $+1$ implies \textit{weak relation}, $0$ to mention \textit{abstain} and $-1$ to denote \textit{no relation} between two tasks $\tau_{i}, \tau_{j} \in \tau$. We use this $\Gamma$ to build our meta-learner \model.}
 \label{fig_gamma}
\end{figure}
\noindent Once we get annotators' error tensor $\mathcal{C}$ and $p(y_j)$ from Equations \ref{eq_dawid_4} and \ref{eq_dawid_5}, we can estimate:
\begin{equation}
    \label{eq_dawid_6}
    p(y_{j} = l|\mathcal{C},\Xi) = \frac{\exp(\sum_{m=1}^{M}\sum_{g=1}^{K}\log\hat{c}_{mlg}\textbf{1}_{(\xi_{m,j}=k)})}{\sum_{l^{'}=1}^{K}\exp(\sum_{m=1}^{M}\sum_{g=1}^{K}\log\hat{c}_{ml^{'}g}\textbf{1}_{(\xi_{m,j}=k)})}
\end{equation}
for all $j \in N$ and $l \in K$. To get the final predicted label, we adopt a winner-takes-all strategy on $p(y_{j}=l)$ across all $l \in K$. We request readers to refer \cite{Zhu2015OnlineC} and \cite{dawid1979maximum} for more details.
\paragraph{Implementation:} In our experiments, as mentioned before, we considered the Taskonomy dataset \cite{zamir2018taskonomy}. This dataset has 26 vision-related tasks. We are interested in finding the task correlation for each pair of tasks in \{$\tau_1, \cdots, \tau_{26}$\}. Let's assume that, we have $M$ annotators. To fit our model in the DS framework, we flatten the task correlation matrix $\Gamma^{26 \times 26}$ (described in section \ref{sec_methodology}) in \textit{row-major} order to get item set $N$ = \{$n_1, \cdots, n_{(26 \times 26)}$\}.
For each item $n_k \in N$, the annotator is asked to give task correlation label on a scale of \{$+3, +2, +1, 0, -1$\}. On that scale, a $+3$ denotes \textit{self relation}, $+2$ describes \textit{strong relation}, $+1$ implies \textit{weak relation}, $0$ to mention \textit{abstain} and $-1$ to denote \textit{no relation} between two tasks $\tau_{i}, \tau_{j} \in \tau$. After getting annotators' vote, we build matrix $\Xi$. Subsequently, we find annotators' error tensor $\mathcal{C}$ (equation \ref{eq_dawid_1}), likelihood estimation (equations \ref{eq_dawid_2}, \ref{eq_dawid_3}, \ref{eq_dawid_4}, \ref{eq_dawid_5}). We get predicted class labels after a \textit{winner-takes-all} in equation \ref{eq_dawid_6}. Predicted class labels are the task correlation we wish to get. We get the final task correlational matrix $\Gamma^{26 \times 26}$, after a de-flattening of $y_{j}^{true}$ for all $j = 1, \cdots, N$. 

Figure \ref{fig_gamma} shows the final $\Gamma$ task correlation matrix used in our experiments. The matrix fairly reflects an intuitive knowledge of the tasks considered. We also considered an alternate mechanism for obtaining the task correlation matrix from the task graph computed in \cite{zamir2018taskonomy}. We present these results later in Section \ref{subsec_task_correlation_ablation}.
\begin{table}[h]
    \centering
    \footnotesize
        \begin{tabular}
            {|c|c|c|c|c|}
            \hline \hline
           \small{Annotators}&\small{RANSAC}\cite{yosinski2014transferable}& LR\cite{korman2018latent}& G3D\cite{zamir2016generic} & TN\cite{zamir2018taskonomy}\\
            \hline \hline
            3& 28\%&22\%&29\%&40\%\\ \hline
            10& 51\%&29\%&31\%&52\%\\ \hline
            20& 90\%&82\%&92\%&42\%\\ \hline
            30& \textit{88}\%&\textit{81}\%&\textit{72}\%&\textit{64}\%\\ \hline
            35& 88\%&82\%&75\%&61\%\\ \hline
            40& 90\%&72\%&69\%&63\%\\ \hline
            45& 87\%&80\%&61\%&70\%\\ \hline
            50& 90\%&82\%&72\%&50\%\\ \hline \hline
        \end{tabular}
    \caption{\textbf{\textit{Win rates (\%)} of $\model_6$ with a varied number of annotators}. We considered the \textit{win rate} (\%) on angular error. Columns are state-of-the-art methods and rows are our $\model_6$ trained using different $\Gamma_i$s, where $i$ = \{$3, 10, 20, 30, 35, 40, 45, 50$\}.}
\label{table_different_annotators}
\end{table}
\begin{figure*}
 \centering 
 \includegraphics[height=11.5cm,width=18cm]{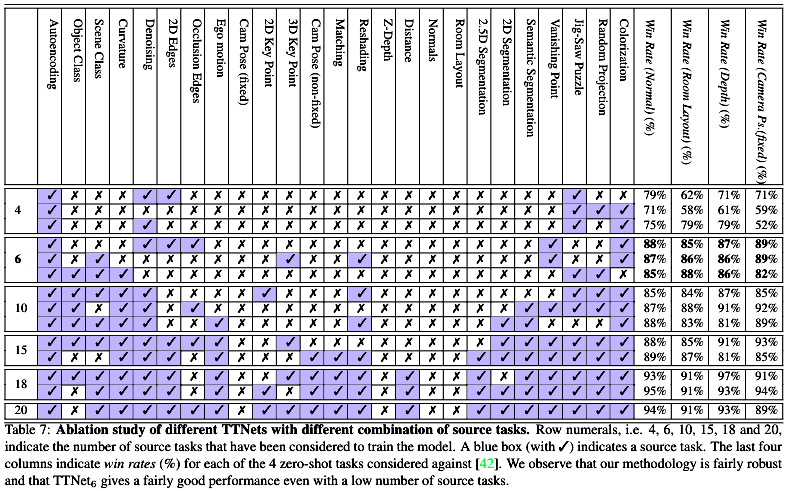}
\end{figure*}
\paragraph{Ablation study on different number of annotators:} The results in the main paper were performed with 30 annotators. In this section, we studied the robustness of our method when $\Gamma$ is obtained by varying the number of annotators, $M_i$, where $i \in \{3, 10, 20, 30, 35, 40, 45, 50\}$.  Table \ref{table_different_annotators} shows a \textit{win rate} (\%) \cite{zamir2018taskonomy} on the camera pose estimation task using \model$_{6}$ (when 6 source tasks are used). While there are variations, we identified $i = 30$ as the number of annotators where the results are most robust, and used this setting for the rest of our experiments (in the main paper). 
\section{Ablation Studies on Varying Known Tasks}
\label{sec_ablation_study_number_of_source_tasks}
In this section, we present two ablation studies w.r.t. known tasks, \textit{viz}, (i) number of known tasks and (ii) choice of known tasks. These studies attempt to answer the questions: how many known tasks are sufficient to adapt to zero-shot tasks in the considered setting? Which known tasks are more favorable to transfer to zero-shot tasks? While an exhaustive study is infeasible, we attempt to answer these questions by conducting a study across six different models: $\text{\model}_{4}$, $\text{\model}_{6}$, $\text{\model}_{10}$, $\text{\model}_{15}$, $\text{\model}_{18}$, and $\text{\model}_{20}$ (where the subscript denotes the number of source tasks considered). 
We used \textit{win rate} (\%) against  \cite{zamir2018taskonomy} for each of the zero-shot tasks. Table 7 shows the results of our studies with varying number and choice of known source tasks. Expectedly, a higher number of known tasks provides improved performance. It is observed from the table that our methodology is fairly robust despite changes in choice of source tasks, and that $\text{\model}_{6}$ provides a good balance by having a good performance even with a low number of source tasks. Interestingly, most of the source tasks considered for $\text{\model}_{6}$ (autoencoding, denoising, 2D edges, occlusion edges, vanishing point, and colorization) are tasks that do not require significant annotation, thus providing a model where 
very little source annotation can help generalize to more complex target tasks on the same domain.
\begin{table}[]
\footnotesize
\begin{tabular}{|p{0.9cm}|p{0.1cm}|p{0.1cm}|p{0.1cm}|p{0.1cm}|p{0.1cm}|p{0.1cm}|p{0.1cm}|p{0.1cm}|p{0.1cm}|p{0.1cm}|p{0.1cm}|p{0.05cm}|}
\hline \hline
\multirow{3}{*}{Model} & \multicolumn{6}{l|}{\text{\model}$_{6}$} & \multicolumn{6}{l|}{Taskonomy} \\ \cline{2-13} 
 & \multicolumn{2}{l|}{Wang} & \multicolumn{2}{l|}{Zamir} & \multicolumn{2}{l|}{Full Sup} & \multicolumn{2}{l|}{Wang} & \multicolumn{2}{l|}{Zamir} & \multicolumn{2}{l|}{Full Sup} \\ \cline{2-13} 
 & N & L & N & L & N & L & N & L & N & L & N & L \\ \hline
Depth & 85 & \textbf{87} & 81 &\textbf{97} & \textbf{67} & 42 & \textbf{98} & 85 & \textbf{92} & 88 & 60 & \textbf{46} \\ \hline
2.5 D & \textbf{88} & 75 & \textbf{75} & \textbf{81} & \textbf{89} & 35 & \textbf{88} & \textbf{77} & 73 & 88 & 85 & \textbf{39} \\ \hline
Curvature & \textbf{84} & 87 & \textbf{91} & 58 & \textbf{86} & 47 & 78 & \textbf{89} & 88 & \textbf{78} & 60 & \textbf{50} \\ \hline \hline
\end{tabular}
\caption{\textbf{Zero-shot to known task transfer.} We consider the autoencoder-decoder parameters for a zero-shot task learned through our method, and finetune the decoder (fixing the encoder) to a target known task, following the procedure in \cite{zamir2018taskonomy}. Source tasks (zero-shot) are surface normal (N), and, room layout (L). Target tasks are depth, 2.5D segmentation and curvature. \textit{Win rates} (\%) of task transfer with respect to self-supervised methods, such as, Wang \textit{et al.} \cite{wang2015designing}, Zamir \textit{et al.} \cite{zamir2016generic} as well as fully supervised setting are shown (all values are in \%), with bold face numerals denoting winning entries.}
\label{table_zero_shot_task_to_known_task_transfer}
\end{table}
\section{Other Results}
\subsection{Zero-shot to Known Task Transfer: Quantitative Evaluation}
In continuation to our discussions in Section \ref{sec_discussion}, we ask ourselves the question: are our regressed model parameters for zero-shot tasks capable of transferring to a known task? To study this, we consider the autoencoder-decoder parameters for a zero-shot task learned through our methodology, and finetune the decoder (fixing the encoder parameters) to a target known task, following the procedure in \cite{zamir2018taskonomy}. Table \ref{table_zero_shot_task_to_known_task_transfer} shows the quantitative results when choosing the source (zero-shot) tasks as surface normal estimation (N) and room layout estimation (L). We compared our \model against \cite{zamir2018taskonomy} quantitatively by studying the \textit{win rate} (\%) of the two methods against other state-of-the-art methods: Wang \textit{et al.} \cite{Wang2017LearningTM}, G3D \cite{zamir2016generic}, and full supervision. However, it is worthy to mention that our parameters are obtained through the proposed zero-shot task transfer, while all other comparative methods are explicitly trained on the dataset for the task.
\begin{figure}[h]
 \centering 
 \includegraphics[width=8cm]{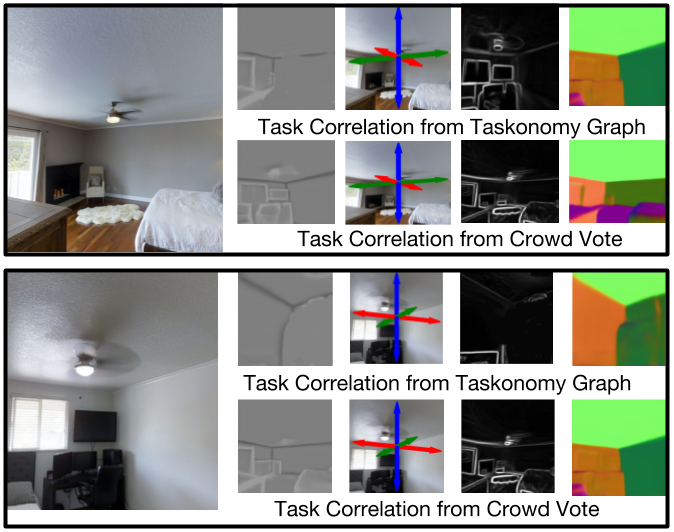}
 \caption{\textbf{Qualitative results of $\text{\model}_{6}$ when task correlation matrix ($\Gamma$) is obtained from task graph computed in \cite{zamir2018taskonomy}.} We studied considering the task graph computed in \cite{zamir2018taskonomy} (instead of crowd vote) to build the task correlation matrix $\Gamma_{TN}$. First column represents RGB image and, subsequent columns (from 2$^{nd}$ to 4$^{th}$ columns) are zero-shot tasks: curvature, vanishing points, 2D key point and surface normal estimation}
 \label{figure_taking_task_graph}
\end{figure}
\subsection{Alternate Methods for Task Correlation Computation}
\label{subsec_task_correlation_ablation}
In our results so far, we studied the effectiveness of computing the task correlation matrix by aggregation of crowd votes. In this section, we instead use the task graph obtained in \cite{zamir2018taskonomy} to obtain the task correlation matrix $\Gamma$. We call this matrix $\Gamma_{TN}$. Figure \ref{figure_taking_task_graph} shows a qualitative comparison of $\text{\model}_{6}$ where the $\Gamma_{TN}$ is obtained from the taskonomy graph, and $\Gamma$ is based on crowd knowledge. It is evident that our method shows promising results on both cases. 

It is worthy to note that although one can use the taskonomy graph to build $\Gamma$: (i) the taskonomy graph is model and data specific \cite{zamir2018taskonomy}; while $\Gamma$ coming from crowd votes does not explicitly assume any model or data and can be easily obtained; (ii) during the process of building the taskonomy graph, an explicit access to zero-shot task ground truth is unavoidable; while, constructing $\Gamma$ from crowd votes is possible without accessing any explicit ground truth.
\begin{figure}
 \centering 
 \includegraphics[height=11cm, width=9cm]{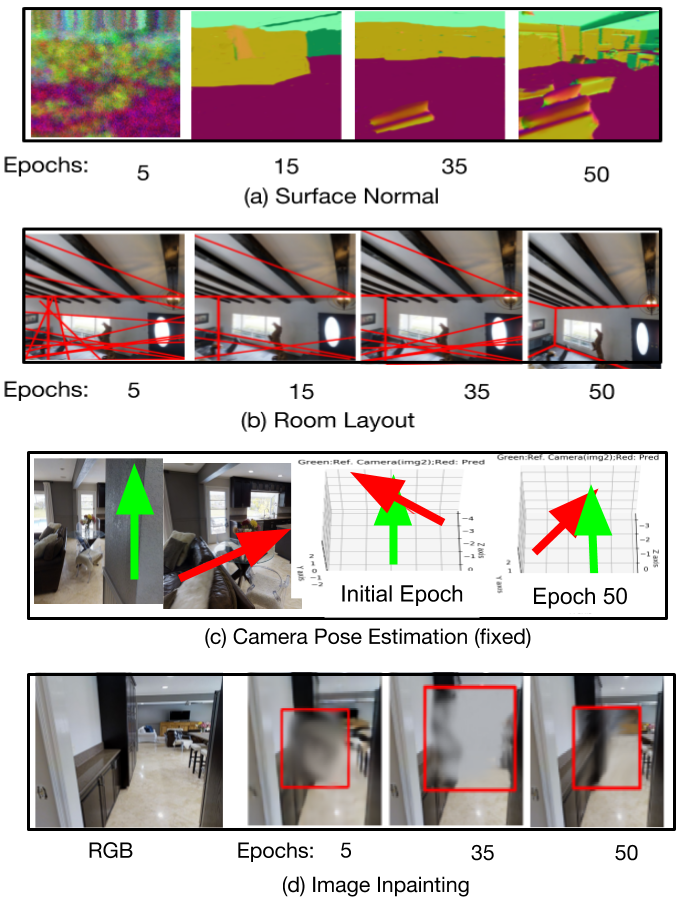}
 \caption{\textbf{Zero-shot tasks results during the training of \model.} We regressed zero-shot task parameters from $\text{\model}_{6}$ during its course of training. The qualitative results show the gradual learning of the model parameters of the epochs.}
 \label{figure_evolution_of_TTNET}
\end{figure}
\subsection{Evolution of \model:} 
Thus far, we showed the final results of our meta-learner after the model is fully trained. We now ask the question - how does the training of the \model model progress over training? We used the zero-shot task model parameters from $\text{\model}_{6}$ during its course of training, and Figure \ref{figure_evolution_of_TTNET} shows qualitative results of different epochs of four zero-shot tasks over the training phase. The results show that the model's training progresses gradually over the epochs, and the model obtains promising results in later epochs. For example, in Figure \ref{figure_evolution_of_TTNET}(a), finer details such as wall boundaries, sofa, chair and other minute details are learned in later epochs.

\subsection{Qualitative Results on Cityscapes Dataset}
\label{sub_Qualitative_Room_Layout}
\begin{figure*}
 \centering 
 \includegraphics[width=17cm]{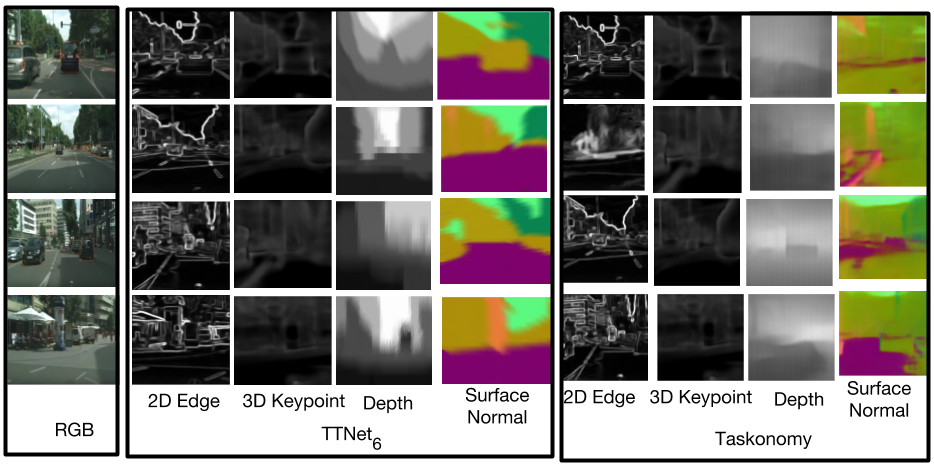}
 \caption{\textbf{Results on Cityscapes data.} We finetuned $\text{\model}_{6}$ on the Cityscapes dataset \cite{Cordts2016Cityscapes}, and the surface normal, depth, 2D edge and 3D keypoint results are reported using the model parameters learned by $\text{\model}_{6}$.}
 \label{fig_city}
\end{figure*}

To further study the generalizability of our models, we finetuned \model on the Cityscapes dataset \cite{Cordts2016Cityscapes}. We get source task model parameters (trained on Taskonomy dataset) to train $\text{\model}_{6}$. We then finetuned $\text{\model}_{6}$ on the segmentation model parameters trained on Cityscapes data. (We modified one source task, i.e. autoencoding to segmentation, of our proposed \model$_{6}$, see table \ref{table_ablation_source_task}, 3$^{rd}$ row. All other source tasks are unaltered.) 
Results of the learned model parameters for four zero-shot tasks, i.e. Surface normal, depth, 2D edge and 3D keypoint, are reported in Figure \ref{fig_city}, with comparison to \cite{zamir2018taskonomy} (which is trained explicitly for these tasks). Despite the lack of supervised learning, the figure shows that tt is evident from the qualitative assessment (figure \ref{fig_city}) that our model seems to capture more detail.
\subsection{More Qualitative Results}
\label{sub_Qualitative_Room_Layout}
\begin{figure*}
 \centering 
 \includegraphics[width=17cm]{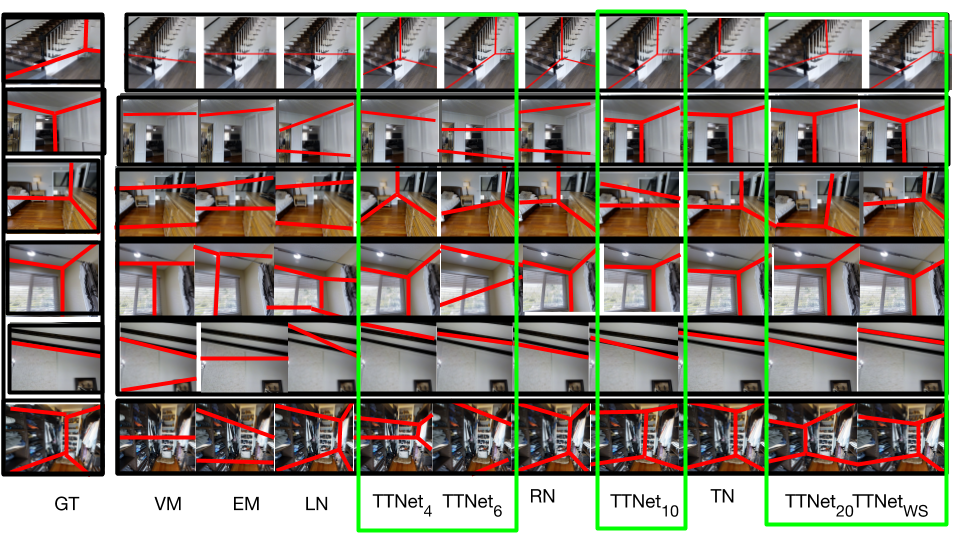}
 \caption{More results of room layout estimation}
 \label{fig_more_room}
\end{figure*}

\begin{figure*}
 \centering 
 \includegraphics[width=17cm]{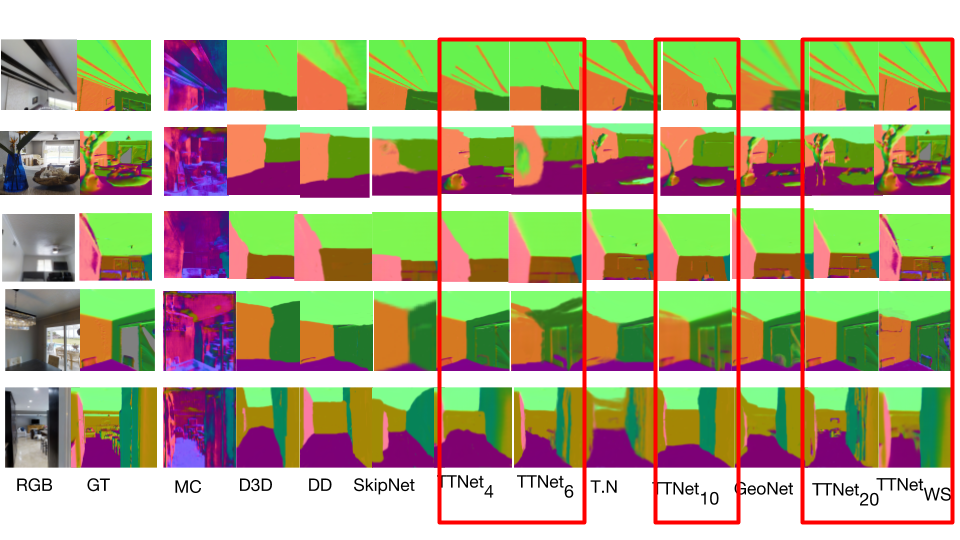}
 \caption{More results of surface normal estimation}
 \label{fig_more_surface}
\end{figure*}

\begin{figure*}
 \centering 
 \includegraphics[width=17cm]{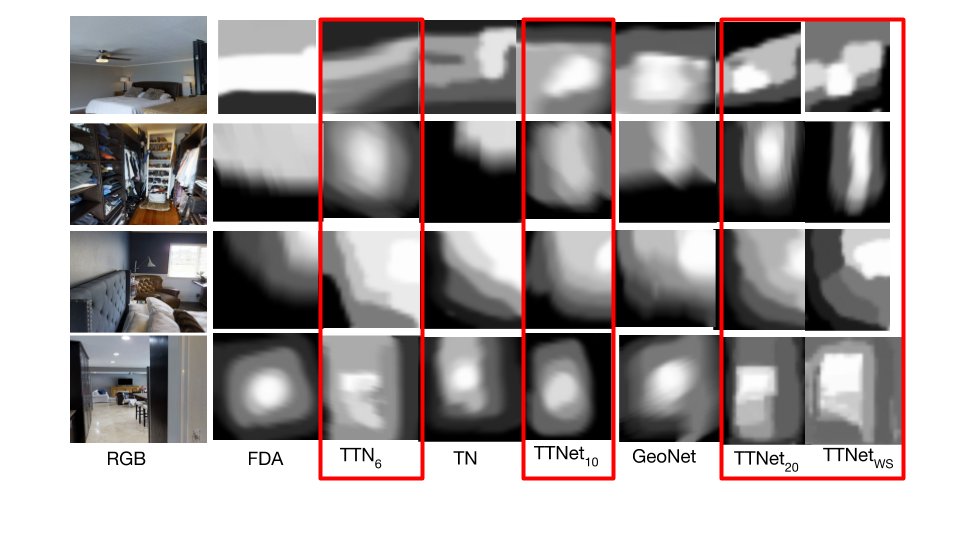}
 \caption{More results of depth estimation}
 \label{fig_more_depth}
\end{figure*}

\begin{figure*}
 \centering 
 \includegraphics[width=17cm]{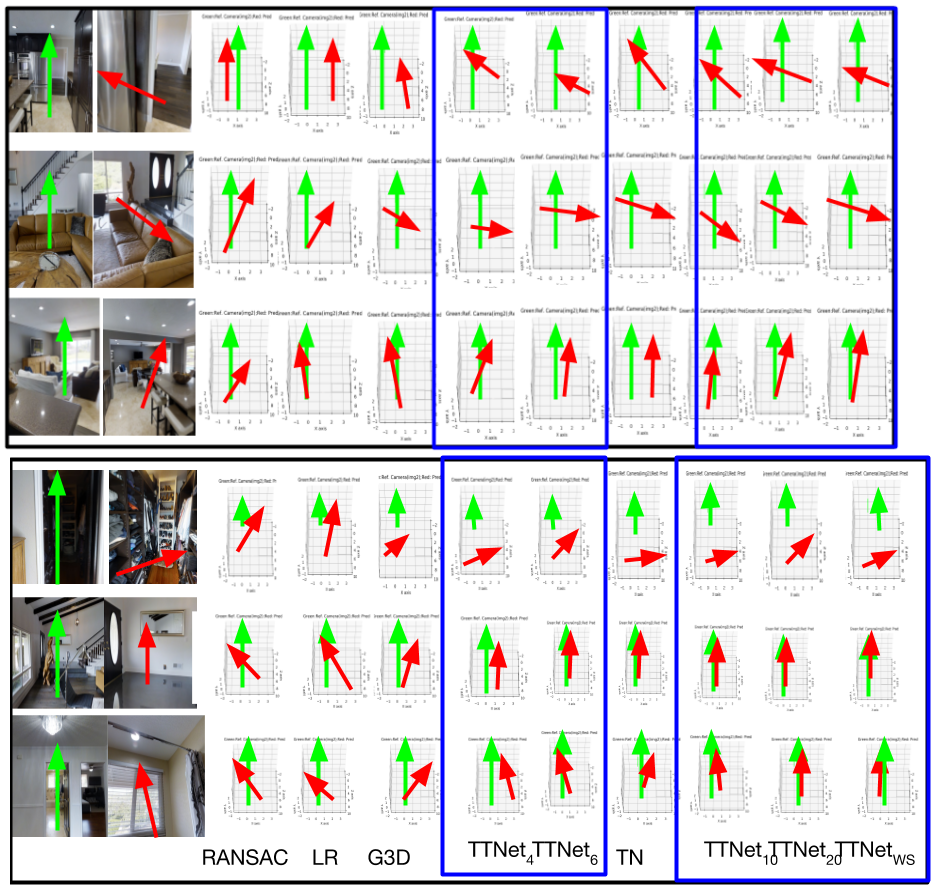}
 \caption{More results of camera pose estimation}
 \label{fig_more_camera}
\end{figure*}

We report more qualitative results of: (i) room layout in Figure \ref{fig_more_room}; (ii) surface normal estimation in Figure \ref{fig_more_surface}; (iii) depth estimation in Figure \ref{fig_more_depth}; and (iv) camera pose estimation in Figure \ref{fig_more_camera}.
\end{document}